\documentclass[letterpaper]{article} % DO NOT CHANGE THIS
\usepackage{aaai2026}  % DO NOT CHANGE THIS
\usepackage{times}  % DO NOT CHANGE THIS
\usepackage{helvet}  % DO NOT CHANGE THIS
\usepackage{courier}  % DO NOT CHANGE THIS
\usepackage[hyphens]{url}  % DO NOT CHANGE THIS
\usepackage{graphicx} % DO NOT CHANGE THIS
\usepackage{natbib}  % DO NOT CHANGE THIS AND DO NOT ADD ANY OPTIONS TO IT
\usepackage{caption} % DO NOT CHANGE THIS AND DO NOT ADD ANY OPTIONS TO IT
\usepackage{algorithm}
\usepackage{algorithmic}

\usepackage{newfloat}
\usepackage{listings}
\DeclareCaptionStyle{ruled}{labelfont=normalfont,labelsep=colon,strut=off} % DO NOT CHANGE THIS
\floatstyle{ruled}
\newfloat{listing}{tb}{lst}{}
\floatname{listing}{Listing}
\usepackage{gensymb}
\usepackage{color}

\title{RaLD: Generating High-Resolution 3D Radar Point Clouds with Latent Diffusion}
\author{
    Ruijie Zhang\textsuperscript{\rm 1}, 
    Bixin Zeng\textsuperscript{\rm 1}, 
    Shengpeng Wang\textsuperscript{\rm 1}, 
    Fuhui Zhou\textsuperscript{\rm 2}, 
    Wei Wang\textsuperscript{\rm 3}\thanks{Corresponding author.}
}
\affiliations{
    \textsuperscript{\rm 1}Huazhong University of Science and Technology, \\
    \textsuperscript{\rm 2}Nanjing University of Aeronautics and Astronautics, 
    \textsuperscript{\rm 3}Wuhan University\\
    \{ruijiezhang, bixinzeng, wsp666\}@hust.edu.cn, zhoufuhui@ieee.org, wangw@whu.edu.cn
}
\nocopyright{}
\usepackage{bibentry}
\usepackage{algorithm}
\usepackage{algorithmic}
\usepackage{amsmath}
\usepackage{amssymb}
\usepackage{multicol}
\usepackage{multirow}
\usepackage{tabularx}
\usepackage{svg}
\usepackage{xcolor}
\usepackage{makecell}
\usepackage{booktabs}
\usepackage{array}
\usepackage{hhline}
\newcommand{\sysname}{RaLD}
\newcommand{\radarspec}{\mathbf{S}}
\newcommand{\pointcloud}{\mathbf{P}}

\begin{document}
\maketitle


\begin{abstract} Millimeter-wave radar offers a promising sensing modality for autonomous
systems thanks to its robustness in adverse conditions and low cost. However, its
utility is significantly limited by the sparsity and low resolution of radar point
clouds, which poses challenges for tasks requiring dense and accurate 3D
perception. 
Despite that recent efforts have shown great potential by exploring generative approaches to address this issue, they often rely on dense voxel representations that are inefficient and struggle to preserve structural detail. To fill this gap, we make the key
observation that latent diffusion models (LDMs), though successful in other modalities,
have not been effectively leveraged for radar-based 3D generation due to a lack
of compatible representations and conditioning strategies. We introduce \sysname{},
a framework that bridges this gap by integrating scene-level frustum-based LiDAR
autoencoding, order-invariant latent representations, and direct radar spectrum
conditioning. These insights lead to a more compact and expressive 
generation process. Experiments show that \sysname{}
produces dense and accurate 3D point clouds from raw radar spectrums, offering a
promising solution for robust perception in challenging environments.
\end{abstract}
\section{Introduction}
Millimeter-wave radar has attracted growing interest in a range of applications,
including perception, localization, and mapping, owing to its affordability, strong
penetration capability, and robustness under adverse weather and unilluminated
conditions~\cite{lu2020see, gao2022dc, fan2024diffusion}. These advantages make
radar an indispensable sensing modality for autonomous systems operating in complex real-world
environments.

However, the utility of radar remains limited by the inherent sparsity and low
resolution of its point cloud outputs, which stem from fundamental hardware constraints.
This poses significant challenges for downstream tasks that require dense and
accurate spatial information. To address this issue, there is a pressing need
for effective methods that can enhance or reconstruct high-quality point clouds
from raw radar measurements, enabling more reliable and precise environmental understanding.

Diffusion models have recently shown remarkable success in cross-modal generation
tasks, such as text-to-image synthesis~\cite{rombach2022high, peebles2023scalable},
inspiring their application in radar-based 3D scene understanding. In the
context of 3D radar point cloud generation, they offer a promising path to generate
high-resolution, LiDAR-like point clouds from sparse and noisy radar
measurements.

\begin{figure}[t]
    \centering
    % \fbox{\rule{0pt}{2.5in} \rule{3in}{0pt}} % Placeholder for a figure
    \includegraphics[width=\linewidth]{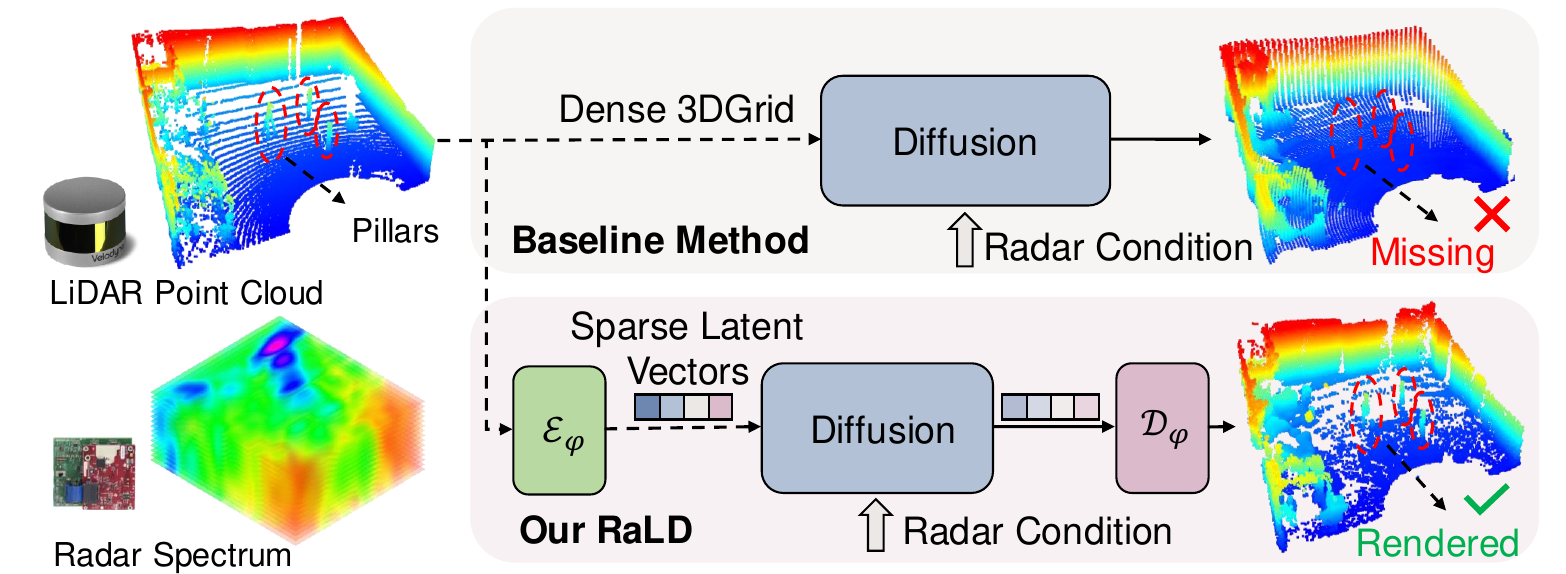}
    \caption{RaLD uses latent diffusion with sparse vectors, enabling high-fidelity 3D radar point cloud generation beyond dense-grid methods.}
    \label{fig-intro}
\end{figure}

Despite their success in image domains, conventional diffusion models struggle
to scale effectively to 3D point clouds, where dense representations incur significant
computational and resolution limitations. While prior work such as SDDiff~\cite{wang2025sddiff}
adopts a dense voxel-based representation for radar-conditioned diffusion, it suffers from high computational cost and memory consumption, which further compromises the
achievable resolution. For example, SDDiff predicts occupancies over a dense
$128\times128\times64$ voxel grid, yet the actual number of generated points remains
on the order of 1\textperthousand—leading to substantial inefficiency and loss of
fine-grained structural details.

In contrast, we observe that latent diffusion models (LDMs)~\cite{rombach2022high}
offer a more efficient and flexible alternative. By operating in a compact latent
space learned via a dedicated autoencoder, LDMs alleviate the burden of modeling
sparse and unordered 3D point clouds directly in data space. This two-stage framework
not only enables higher-resolution generation with reduced resource demands but also
better preserves structural details, making it well-suited for the challenges of
radar-based 3D point cloud super-resolution.

While latent diffusion offers a promising direction, applying it to radar-conditioned
3D point cloud generation poses several technical challenges. First, a key requirement
is a robust autoencoder that can compress high-resolution, scene-level LiDAR
point clouds into informative latent codes. Most existing point cloud autoencoders
are designed for object-level shapes~\cite{zhang20233dshape2vecset} and fall short
in handling the sparse, large-scale, and geometrically complex nature of LiDAR data.
Second, raw radar spectrums are extremely noisy and lack strong semantic
cues, especially compared to LiDAR. These spectrums are affected by multipath effects,
specularity, resulting in ambiguous or aliased observations. Conditioning
generation on such data is non-trivial, as naive approaches (e.g., radar point-based
inputs) often fail to provide sufficient guidance for high-quality reconstruction.

To overcome these challenges, we propose a set of complementary designs that
enable effective radar-to-LiDAR point cloud generation within the latent
diffusion framework. We first introduce a point-based LiDAR autoencoder that compresses
sparse scene-level point clouds into compact latent vectors and reconstructs them
by predicting occupancy at arbitrary 3D query locations. Instead of using
Cartesian voxel grids, we define occupancy labels based on volumetric frustums aligned
with the LiDAR's polar sampling geometry. This approach preserves the spatial
structure of LiDAR returns, captures their non-uniform density across depth, and
enables physically grounded interpolation during decoding. Building on this, we
design an order-invariant latent encoding strategy that fuses local and global
geometric features into a hybrid token representation, ensuring the latent codes
are both expressive and order-invariant—crucial properties for stable and generalizable
diffusion modeling. Finally, instead of conditioning on sparse radar point clouds,
we leverage the richer, albeit noisy, radar spectrum. Our radar spectrum guidance
module extracts spatial cues from the signal, providing strong
priors for generation and improving the robustness and efficiency of decoding in
sparse or ambiguous environments.

Through this integrated design, \sysname{} enables end-to-end generation of
dense 3D structures from raw radar signals, offering a practical and scalable solution
for robust 3D perception in adverse environments. In summary, this paper makes
the following contributions:
\begin{itemize}
    \item To our knowledge, this is the first study to explore a sparse, point-based
        latent diffusion model for radar spectrum-conditioned 3D point cloud generation,
        effectively enhancing the fidelity and structural quality of point
        clouds generated from raw radar spectrums.

    \item We develop a set of complementary designs tailored for radar-conditioned
        latent diffusion, including a frustum-based LiDAR autoencoder that
        preserves the polar geometry of LiDAR scans, an order-invariant latent encoding
        strategy that fuses local and global geometric features, and a radar
        spectrum guidance module that extracts semantic and spatial cues from
        raw radar signals to guide generation. Together, these components enable
        robust and high-resolution 3D point cloud synthesis from sparse and
        ambiguous radar inputs.

    \item Compressive experiments on ColoRadar dataset demonstrate that our method
        significantly outperforms existing methods in terms of point cloud quality,
        achieving state-of-the-art results in radar 3D point cloud generation.
\end{itemize}
\section{Related Work}
\subsubsection{Traditional Radar Super-Resolution.}
Extracting dense and reliable point clouds from low-resolution radar data has
been a long-standing challenge in the field of radar signal processing. Traditional
methods for radar super-resolution often rely on signal processing techniques,
such as MUSIC~\shortcite{schmidt1986multiple}, and ESPRIT~\shortcite{roy1990esprit},
collaborating with a Const Fasle Alarm Rate (CFAR) detection strategy~\cite{richards2005fundamentals}
to generate point clouds from radar spectrums. More recent approaches ~\cite{qian20203d,
lai2024enabling, geng2024dream}
have focused on leveraging synthetic aperture radar (SAR) techniques ~\cite{soumekh1999synthetic}
to enhance the resolution of radar images. However, SAR-based methods typically rely
on predefined moving trajectories and precise estimation of the platform's motion,
which limits their application scenarios.

\subsubsection{Radar Super-Resolution with Generative Models.}
% Early attempts~\cite{guan2020through, sun20213drimr, cheng2022novel, fan2024enhancing}
% to apply generative models in radar super-resolution primarily focused on using
% Generative Adversarial Networks (GANs) ~\cite{goodfellow2020generative}. These
% works typically employed accumulated radar spectrums to generate high-resolution
% point clouds~\cite{guan2020through, cheng2022novel, prabhakara2022high}. However,
% GANs suffer from notoriously unstable training behavior~\cite{kurach2019large}, which
% hampers their reliability. With the remarkable success of diffusion models~\cite{ho2020denoising},
% researchers have begun exploring their applications in radar super-resolution.
Early works~\cite{guan2020through, sun20213drimr, cheng2022novel, fan2024enhancing}
on radar super-resolution mainly used GANs~\cite{goodfellow2020generative} with
accumulated radar spectrums to generate high-resolution point clouds~\cite{guan2020through,
cheng2022novel, prabhakara2022high}. However, GANs suffer from unstable training~\cite{kurach2019large},
limiting their reliability. Recent success of diffusion models~\cite{ho2020denoising}
has sparked interest in their use for this task. These methods typically use high-quality
LiDAR point clouds as ground truth and train generative models that map radar data
to LiDAR-like outputs. Zhang et al.~\shortcite{zhang2024towards} first
introduced diffusion models for radar super-resolution, proposing an efficient framework
to generate high-resolution point clouds in edge devices with limited computational
resources. However, it is limited to generating point clouds in 2D, which leads
to a significant loss of information in the 3D space. Luan et al.~\shortcite{luan2024diffusion}
further extended to 3D point cloud generation model in bird's-eye view (BEV) space,
though BEV still poorly captures hights geometry. SDDiff ~\cite{wang2025sddiff}
proposed a directional diffusion model, which starts from the radar prior distribution
and diffuses toward the 3D LiDAR point clouds. Although both Luan et al.~\shortcite{luan2024diffusion}
and SDDiff~\shortcite{wang2025sddiff} are capable of generating 3D point clouds,
their resolution remains limited. This is largely due to their reliance on dense
representations—such as BEV images or 3D voxel grids—as the generation target,
which are computationally expensive and restrict output fidelity. In contrast,
our method proposes a latent diffusion model that operates in a lower-dimensional
latent space, significantly improving computational efficiency and enabling the generation
of high-resolution 3D point clouds.
\begin{figure*}[t]
    \centering
    % \fbox{\rule{0pt}{3in} \rule{7in}{0pt}} % Placeholder for a figure
    \includegraphics[width=\linewidth]{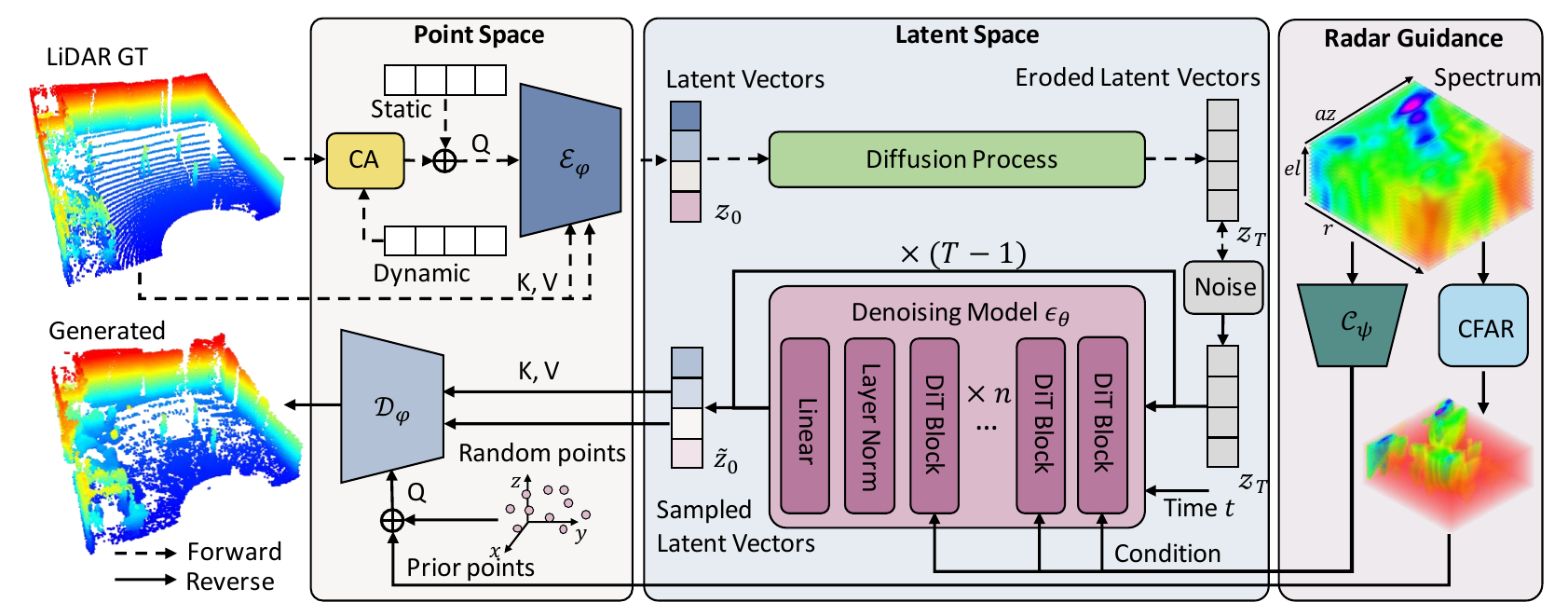}
    \caption{Framework of \sysname~pipeline.}
    \label{fig-system-overview}
\end{figure*}
\subsubsection{3D Latent Diffusion Models.}
Vision generation has been revolutionized by diffusion models, which have shown remarkable
performance in generating high-quality images~\cite{ho2020denoising}. Stable
Diffusion~\cite{rombach2022high} has further advanced the field by introducing a
latent diffusion model that operates in a lower-dimensional latent space, significantly
improving computational efficiency and enabling the generation of high-resolution
images. Following the success of 2D latent diffusion models, researchers have explored
their application in 3D content generation. 3DShape2Vectset~\cite{zhang20233dshape2vecset}
introduced an effective representation for object-level 3D shapes in the latent
space of 1D vectors, which serves as a foundation 3D variational autoencoder (VAE)
for the folloing 3D latent diffusion models ~\cite{zhao2023michelangelo, zhang2024clay,
li2024craftsman3d}. However, these methods primarily focus on object-level 3D
generation, which limits their applicability in generating scene-level 3D content.
Moreover, these models are designed to generate 3D shapes on condition of text or
images, leaving an unexplored gap in generating 3D LiDAR-like point clouds
conditioned on radar spectrums.

\section{Methodology}
In this section, we first provide a brief overview of latent diffusion models as
a preliminary. Then, we present the overall system architecture of our proposed method,
followed by design details of the frustum-based LiDAR autoencoder, 
the order-invariant latent encoding, and the radar spectrum guidance.

\subsection{Preliminary}
\subsubsection{Diffusion Models}
Diffusion models are a class of generative models that learn to generate data by
reversing a diffusion process~\cite{ho2020denoising}. The process begins with a simple
distribution, such as Gaussian noise, and gradually transforms it into a complex
data distribution through a series of steps.

The forward diffusion process is defined as a Markov chain that adds Gaussian noise
to the data at each step:
\begin{equation}
    q(x_{t}\mid x_{t-1}) = \mathcal{N}(x_{t}; \sqrt{1 - \beta_{t}}x_{t-1}, \beta_{t}
    \mathbf{I}), t=1,\cdots T
\end{equation}
where $x_{0}$ is the original data, $x_{T}$ is the final noisy data, and
$\beta_{t}$ controls the noise schedule. 
The reverse process is parameterized by a neural network and aims to recover the
original data by removing noise step by step:
\begin{equation}
    p_{\theta}(x_{t-1}\mid x_{t}, c) = \mathcal{N}(x_{t-1}; \mu_{\theta}(x_{t}, t
    , c), \Sigma_{\theta}(x_{t}, t, c)),
\end{equation}
where $\mu_{\theta}$ and $\Sigma_{\theta}$ are the mean and variance predicted by
the network, and $c$ denotes optional conditioning information, such as labels,
text, or radar signals. The model is trained to predict the added noise, typically
using a mean squared error loss.

\subsubsection{Latent Diffusion Models}
Latent Diffusion Models (LDMs)~\cite{rombach2022high} extend diffusion models by
operating in a lower-dimensional latent space instead of directly in the data space.
For example, a pre-trained autoencoder
$(\mathcal{E}_{\boldsymbol{\varphi}}, \mathcal{D}_{\boldsymbol{\varphi}})$ is
used to compress high-dimensional point cloud data $x \in \mathbb{R}^{N\times3}$ into a compact
latent representation $z\in \mathbb{R}^{M\times d}$, where $N$ is the number of points,
$d$ is the dimension of channel, and $M$ is the number of latent points (typically $M \ll N$).:
\begin{equation}
    z = \mathcal{E}_{\boldsymbol{\varphi}}(x), \quad x \approx \mathcal{D}_{\boldsymbol{\varphi}}
    (z),
\end{equation}
where $\mathcal{E}_{\boldsymbol{\varphi}}$ and $\mathcal{D}_{\boldsymbol{\varphi}}$
denote the encoder and decoder, respectively.

The diffusion model is then trained in the latent space by perturbing $z$ and
learning to reverse the corruption:
\begin{equation}
    z_{t}= \sqrt{\bar{\alpha}_{t}}z_{0} + \sqrt{1 - \bar{\alpha}_{t}}\epsilon, \quad
    \epsilon \sim \mathcal{N}(0, \mathbf{I}),
\end{equation}
\begin{equation}
    \mathcal{L}_{\text{LDM}}= \mathbb{E}_{z_{t}, \epsilon, t}\left[\left\| \epsilon -
    \epsilon_{\theta}(z_{t}, t) \right\|^{2}\right].
\end{equation}
Once denoised, the final latent $z_{0}$ is decoded to reconstruct the data $x \approx
\mathcal{D}(z_{0})$.

\subsection{System Overview}
Given an input radar spectrum $\radarspec{}$, \sysname{} aims to generate a dense
and accurate 3D point clouds $\mathbf{P}\in \mathbb{R}^{N \times 3}$ that reconstruct
the scene with LiDAR-like fidelity. Following prior works~\cite{zhang2024towards,
wang2025sddiff}, we adopt a conditional diffusion framework that learns to synthesize
point clouds conditioned on radar observations.

To achieve this, \sysname{} operates in a compact latent space, where a
diffusion model is trained to generate point cloud embeddings guided by the
radar spectrum. The overall pipeline, as illustrated in Figure~\ref{fig-system-overview},
begins with an autoencoder that compresses LiDAR point clouds into structured latent
codes. A radar-conditioned latent diffusion model then samples from this space, and
a decoder reconstructs the final 3D point clouds guided by radar priors.

We next present the key designs that collectively realize \sysname{}'s capability:
a frustum-based LiDAR autoencoder, an order-invariant latent representation, and
a radar-aware generation strategy. Together, these components form a cohesive
system that maps radar signals to detailed LiDAR-like point clouds.

\subsection{Frustum-Based LiDAR Autoencoder}
While latent diffusion enables efficient generation of 3D point clouds, designing
an effective autoencoder remains challenging due to the extreme sparsity and irregular
structure of scene-level LiDAR data. To address this, we propose a tailored autoencoder
architecture that leverages the geometric characteristics of LiDAR point clouds.

Inspired by prior vector set-based representations for 3D shapes~(\citeauthor{zhang20233dshape2vecset}),
we compress scene-level LiDAR point clouds into a set of compact latent vectors,
and reconstruct them by predicting the occupancy of query points in space. By treating
the decoder as a continuous interpolation function for occupancy prediction,
this design allows high-fidelity reconstruction with a flexible number of output points.

A key challenge lies in defining the occupancy label for each query point, as required
by the decoder. A common approach, widely adopted in perception tasks such as 3D
object detection~\cite{zhou2018voxelnet, yin2021center}, defines occupancy and extracts
features using voxel grids in Cartesian space, where a query point is considered
occupied if its voxel contains at least one LiDAR return. However, it fails to
capture the sensing characteristics of LiDAR. In particular, LiDAR sensors emit
beams at fixed angular resolutions, resulting in a non-uniform point distribution-dense
in the near field and sparse at greater distances. Cartesian voxelization
ignores this property, imposing a uniform spatial grid that does not align with how
the data is captured.

\begin{figure}[tb]
    \centering
    % \fbox{\rule{0pt}{2.5in} \rule{3in}{0pt}} % Placeholder for a figure
    \includegraphics[width=\linewidth]{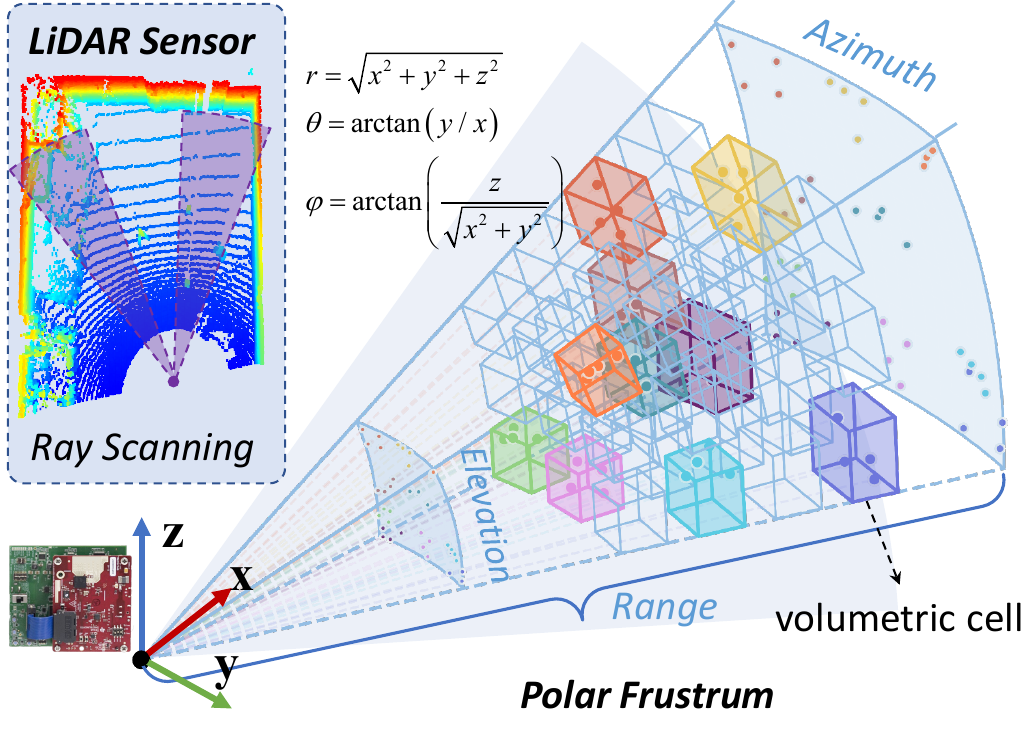}
    \caption{Frustum-based occupancy partitioning aligned with the capture geometry of LiDAR and radar. Volumetric cells are defined in polar coordinates—range, azimuth, and elevation—to match the sensors’ angular sampling patterns.
}
    \label{fig-design-frustum}
\end{figure}

To better align with LiDAR sensing geometry, we define occupancy using frustums
in polar coordinates as shown in Figure~\ref{fig-design-frustum}. These grids respect the angular
sampling pattern of the sensor and preserve spatial regularity across depth.
Formally, a frustum $\mathcal{F}_{i,j,k}$ is defined as a volumetric cell in polar
coordinate space, bounded by range, azimuth, and elevation intervals. Specifically,
the frustum is defined as:
\begin{equation}
    \mathcal{F}_{i,j,k}= \left\{ (r, \theta, \phi) \ \middle|
    \begin{aligned}
        r      & \in [r_{i}, r_{i+1}) \subseteq [r_{\min}, r_{\max}]                     \\
        \theta & \in [\theta_{j}, \theta_{j+1}) \subseteq [\theta_{\min}, \theta_{\max}] \\
        \phi   & \in [\phi_{k}, \phi_{k+1}) \subseteq [\phi_{\min}, \phi_{\max}]
    \end{aligned}
    \right\},
\end{equation}
where $r$ is the radial distance, $\theta$ is the azimuth angle, and $\phi$ is the
elevation angle. The intervals $[r_{\min}, r_{\max}]$, $[\theta_{\min}, \theta_{\max}
]$, and $[\phi_{\min}, \phi_{\max}]$ denote the LiDAR's range, azimuth, and
elevation field of view, respectively. Each frustum $\mathcal{F}_{i,j,k}$ thus
captures a local volume along a LiDAR ray direction.

We define the occupancy $O$ of a frustum $\mathcal{F}_{i,j,k}$ as a binary
indicator:
\begin{equation}
    O_{i,j,k}=
    \begin{cases}
        1, & \text{if }\exists\, \mathbf{p}\in \mathcal{P}\text{ such that }\mathbf{p}\in \mathcal{F}_{i,j,k} \\
        0, & \text{otherwise}
    \end{cases},
\end{equation}
where $\mathcal{P}$ denotes the set of all LiDAR points in the scene. The
occupancy of a query point $\mathbf{q}$ is then determined by the occupancy of
the frustum $\mathcal{F}_{i,j,k}$ that contains it:
\begin{equation}
    O(\mathbf{q}) = O_{i,j,k}\quad \text{if }\mathbf{q}\in \mathcal{F}_{i,j,k}.
\end{equation}

Critically, frustum-based partitioning also facilitates learning occlusion relationships,
since occupied frustums tend to be the closest along the same angular ray. This geometric
alignment makes the decoder's task of occupancy prediction, which can be viewed
as an interpolation over 3D space, more structured and physically grounded. Besides, partinioning
the space into frustums in polar coordinates provides a more consitistent representation for radar
spectrum, which makes it easier to utilize the radar spectrum as the conditioning
information for the diffusion model.

% Our design is inspired by VecSet~\cite{zhang20233dshape2vecset}, a foundational
% framework for object-level 3D shape representation. VecSet encodes 3D shapes into
% a set of compact 1D latent vectors using a cross-attention-based encoder, and
% reconstructs shapes by predicting the occupancy of query points through a
% learned decoder. This design allows high-fidelity reconstruction with a flexible
% number of output points, treating the decoder as a continuous shape
% interpolation function.

% We extend this idea to the scene level by constructing a LiDAR autoencoder that
% compresses sparse 3D point clouds into a compact latent set. A central challenge
% is how to define the occupancy label of each query point, as required by the
% decoder. In VecSet, occupancy is determined by whether a point lies inside a
% solid shape. However, LiDAR point clouds consist only of surface returns, meaning
% points that lie on visible surfaces, which calls for a different approach to
% occupancy labeling.

\subsection{Order-Invariant Latent Encoding}

While the frustum-based autoencoder provides a compact latent representation aligned with LiDAR sensing geometry, it is also essential to ensure that the learned
latent space respects the unordered nature of point clouds. To this end, we
introduce an enhanced order-invariant encoding mechanism that improves the
generalization and stability of the latent diffusion process.

\begin{figure}[t]
    \centering
    % \fbox{\rule{0pt}{2.5in} \rule{3in}{0pt}} % Placeholder for a figure
    \includegraphics[width=\linewidth]{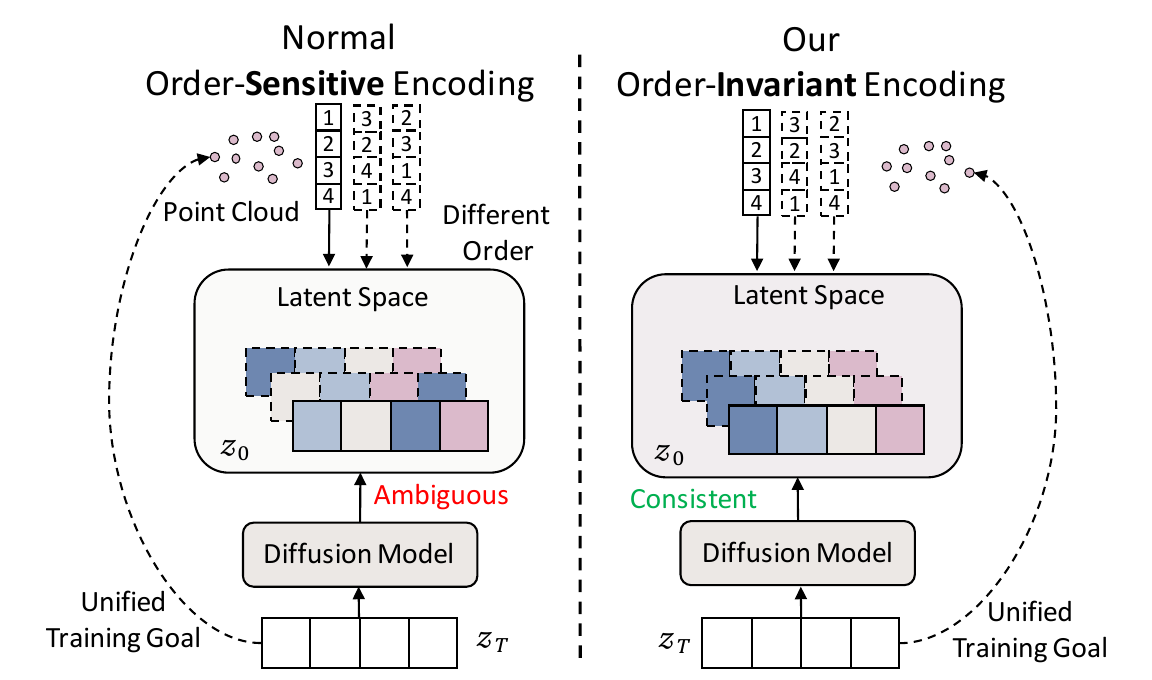}
    \caption{Order-sensitive (left) vs. order-invariant (right) latent encodings.
    Our method ensures consistent training targets despite the order of input point cloud.}
    \label{fig-order-invariant}
\end{figure}

Point clouds are unordered sets, meaning any permutation of points in {$\mathbf{P}$}
should represent the same geometry. However, neural encoders may inadvertently produce
order-sensitive representations, which become problematic in diffusion training.
During training, diffusion models are optimized to predict the noise added to
latent variables at each timestep. If the encoded latent tokens change under different
point orderings, then the target noise also varies inconsistently—even when the
underlying geometry remains the same. This mismatch alters the optimization trajectory
for the same training sample, impairing the model's ability to learn a stable
and generalizable denoising function. We visualize this issue in Figure~\ref{fig-order-invariant},
which highlights how different input orderings lead to inconsistent noise
prediction targets.

To address this, we design a token encoding scheme that ensures consistent latent
representations regardless of input point order. The key idea is to control the
tokens fed into the cross-attention-based encoder. Rather than adopting either randomly
sampled or fixed learned queries, as done in prior work~\cite{zhang20233dshape2vecset},
we employ a \textit{hybrid strategy} that integrates both static and dynamic
queries.

Concretely, a fixed set of learned tokens serves as stable anchors, ensuring
consistent token ordering across different samples. In parallel, dynamic queries
are derived from {$\mathbf{P}$} via a learnable projection, allowing the encoder
to capture geometry-specific features through cross-attention. The fusion of
static and dynamic components forms the final set of query tokens for encoding.

Formally, we denote $\mathbf{Q}_{\text{s}}\in \mathbb{R}^{M \times d}$ as a fixed
set of learnable query tokens that provide consistent ordering across samples.
In parallel, dynamic queries $\mathbf{Q}_{\text{d}}$ are computed from {$\mathbf{P}$}
to capture input-specific geometric features. The final encoder queries
$\mathbf{Q}_{enc}$ are obtained by applying cross-attention between the dynamic
queries and the input, followed by combining the result with the static queries:
\begin{equation}
    \mathbf{Q}_{enc}= \text{Proj}(\mathbf{Q}_{\text{s}}+ \text{CrossAttn}(\mathbf{Q}
    _{\text{d}}, \pointcloud{})).
\end{equation}

This hybrid design maintains order invariance through fixed query structure,
while enriching the latent space with geometry-aware features. It thus enables more
effective diffusion modeling on unordered point clouds.

\subsection{Radar Spectrum Guidance}
\subsubsection{Diffusion Conditioned on Radar Spectrum}
To incorporate radar observations into the generation process, we propose a radar-guided
conditioning strategy that injects semantic and geometric cues from the radar spectrum
into the latent diffusion model. Radar measurements are represented as a 3D
tensor in polar coordinates (range, azimuth, elevation), where high-intensity values
correspond to strong surface reflections. However, the radar signal is inherently
noisy and exhibits anisotropic uncertainty—range measurements tend to be more reliable
than angular ones due to beam characteristics~\cite{kramer2022coloradar}.

To handle this, we upsample the radar spectrum along the azimuth and elevation axes
and apply a convolutional encoder $\mathcal{C_{\psi}}$ to extract robust high-level features. This design
enlarges the receptive field, suppressing angular noise while emphasizing more
reliable range information. The resulting feature map serves as a compact and noise-aware
representation of the radar input.

Since the latent space is formulated as 1D vectors, we adopt the transformer-based
DiT architecture~\cite{peebles2023scalable} as our diffusion model. The radar features
are enriched with 3D positional embeddings to preserve spatial structure, then
flattened to match the latent vector shape. This enables the diffusion model to capture
spatially-aware relationships between the latent tokens and the radar signal.

\subsubsection{Decoding with Radar-Guided Query Initialization}
After sampling from the latent diffusion model, we decode the resulting latent representations
into point clouds. Due to the extreme sparsity of scene-level point clouds, this
decoding process is computationally challenging. Although our autoencoder is
trained with high-resolution frustums (centimeter-level in range and sub-degree in
angle), it is infeasible to densely query the entire 3D space during decoding,
as this would require evaluating tens of millions of points.

To improve efficiency, we introduce a radar-guided query initialization strategy.
Specifically, we apply a low-threshold CFAR algorithm
to the radar spectrum to identify candidate object regions. These CFAR
detections guide the selection of query points for decoding. While imperfect,
they provide a strong prior for object locations and reduce unnecessary queries
in empty space. To maintain completeness, we additionally include a set of
randomly sampled query points across the 3D space, allowing the decoder to capture
undetected or low-reflectivity structures.
\section{Experiments}

\subsection{Experiment Settings}
\subsubsection{Dataset}
We evaluate our method on ColoRadar~\cite{kramer2022coloradar} dataset, which provides
radar spectrum data paired with LiDAR point clouds. The dataset is collected in
various scenarios, including laboratories, hallways, and other environments.
Each environment contains multiple sequences. We use the earlier sequences for training
and reserve the last two sequences in each scene for validation and testing. To
align radar and LiDAR data, we remove non-overlapping frames, transform LiDAR
point clouds using the provided calibration parameters, and crop them to match the
radar's field of view and finally convert them to polar coordinates.

\subsubsection{Implementation Details}
For the autoencoder, the occupancy granularity is set to a frustum with
0.05 m, 0.25°, and 0.5° resolution for the range, azimuth, and elevation, respectively.
Input point cloud coordinates are normalized to [-1,1]. During training, point clouds
are downsampled to 10,000 points, with an equal number of decoder query points. To
address sparsity, 6.25\% of queries are positives; the rest are randomly sampled
negatives. Points are encoded and compressed into 512 latent tokens, each with
32 dimensions. The autoencoder is trained for 150 epochs with a batch size of 28.

The diffusion model is trained for 100 epochs with a batch size of 16 using the
EDM sampler~\cite{Karras2022edm}. At inference, 500k query points are sampled
from free space and 700k from CFAR regions to guide generation. All models are
implemented in PyTorch and trained on two NVIDIA RTX 4090 GPUs. Training takes 28
hours for the autoencoder and 60 hours for the diffusion model.

\subsubsection{Metrics and Baselines}
We evaluate Chamfer distance (CD) and Earth Mover's Distance (EMD) to measure the
similarity between the generated point cloud and the ground truth, both for autoencoder
and diffusion model outputs. CD measures the average distance between points in the
generated and ground truth point clouds, while EMD quantifies the minimum cost
of transforming one point cloud into another, considering the distribution of points.
Lower values indicate better performance. We compare our method with the traditional
method OS-CFAR~\cite{richards2005fundamentals}, as well as the GAN-based method
RPDNet~\cite{cheng2022novel} and diffusion-based method SDDiff~\cite{wang2025sddiff}.

\subsection{Results}
\subsubsection{Main Results}
\begin{table}[b]
    \centering
    \fontsize{9pt}{10pt}\selectfont
    \setlength{\tabcolsep}{4pt} % control hor spacing
    \begin{tabular}{@{}lccccc@{}}
        \hhline{======} \toprule \multirow{2}{*}{\textbf{Scene}} & \makecell[c]{\textbf{Encoder}\\\textbf{Query}} & \textbf{Hybrid} & \textbf{Sample} & \textbf{Static} & \textbf{Hybrid} \\
        \cmidrule{2-6}                                           & \textbf{Occupancy}                             & Voxel           & Frustum         & Frustum         & Frustum         \\
        \midrule \multirow{2}{*}{Aspen Lab}                      & CD↓                                            & 0.133           & \textbf{0.082}  & 0.090           & 0.088           \\
                                                                 & EMD↓                                           & 0.132           & \textbf{0.083}  & 0.089           & 0.087           \\
        \midrule \multirow{2}{*}{Hallways}                       & CD↓                                            & 0.166           & \textbf{0.094}  & 0.118           & 0.112           \\
                                                                 & EMD↓                                           & 0.162           & \textbf{0.095}  & 0.118           & 0.112           \\
        \midrule \multirow{2}{*}{ARPG Lab}                       & CD↓                                            & 0.160           & 0.082           & 0.104           & \textbf{0.081}  \\
                                                                 & EMD↓                                           & 0.155           & 0.083           & 0.104           & \textbf{0.080}  \\
        \bottomrule \hhline{======}
    \end{tabular}
    \caption{Performance comparison of different autoencoders.}
    \label{tab:main-ae}
\end{table}
We first present the main results of the proposed autoencoder in Table~\ref{tab:main-ae}.
We compare the performance of different autoencoders across three scenes: Aspen
Lab, Hallways, and ARPG Lab. To validate the effectiveness of the method of occupancy
partitioning, we compare the performance of the hybrid query with the voxel-based
and the frustum-based partitioning. And we also compare the performance of the hybrid
query with the static query and downsampled point query in the frustum-based partitioning.
All the results are obtained using 500k randomly sampled points from the free
space, with the voxel size in the voxel-based query set to [0.05, 0.05, 0.05] meters
along the x, y, and z axes for fair comparison.

The results show that the frustum-based occupancy partitioning strategy outperforms
the voxel-based counterpart with a significant margin, which provides a maximum
decrease of 49.6\% in CD and 48.3\% in EMD. Within the frustum-based setting,
the hybrid query consistently outperforms the static query across all scenes and
achieves performance comparable to the downsampled point query. The results indicate
the hybrid query in the frustum-based partitioning can provide a higher upper bound
of the downstream latent diffusion task, while preserving the characteristics of
order-invariant encoding.
\begin{figure*}[ht]
    \centering
    % \fbox{\rule{0pt}{3in} \rule{7in}{0pt}} % Placeholder for a figure
    % \includegraphics[width=\linewidth]{figs/visualize.pdf}
    \includegraphics[width=\linewidth]{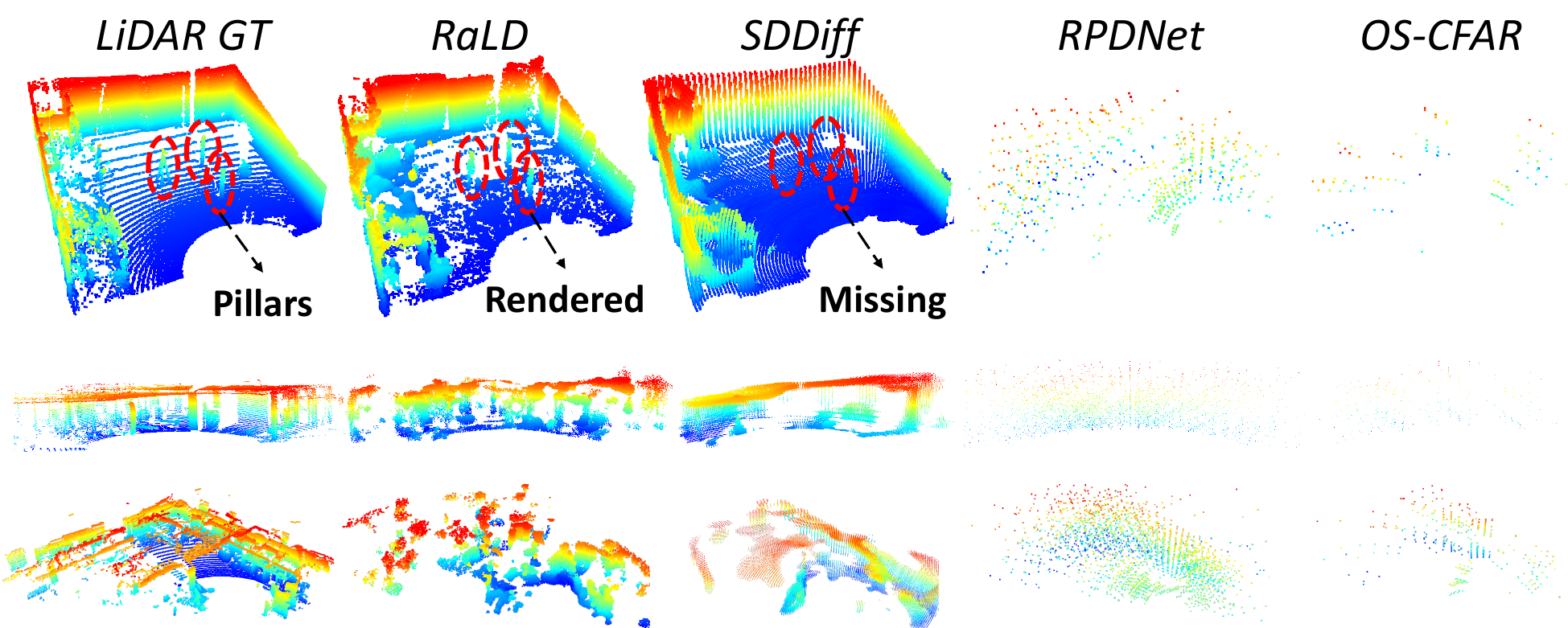}
    \caption{Visualization of 3D radar point cloud generation result.}
    \label{fig-vis-result}
\end{figure*}

\begin{table}[t]
    \centering
    \fontsize{9pt}{10pt}\selectfont
    \setlength{\tabcolsep}{4pt}
    \begin{tabular}{lcccccc}
        \hhline{=======} \toprule \multirow{2}{*}{\textbf{Model}} & \multicolumn{2}{c}{\textbf{Aspen Lab}} & \multicolumn{2}{c}{\textbf{Hallways}} & \multicolumn{2}{c}{\textbf{ARPG Lab}} \\
        \cmidrule(lr){2-3} \cmidrule(lr){4-5} \cmidrule(lr){6-7}  & CD↓                                    & EMD↓                                  & CD↓                                  & EMD↓           & CD↓            & EMD↓           \\
        \midrule OS-CFAR                                          & 1.175                                  & 1.342                                 & 1.098                                & 1.387          & 1.076          & 1.163          \\
        RPDNet                                                    & 0.874                                  & 0.587                                 & 0.793                                & 0.664          & 0.823          & 0.512          \\
        SDDiff                                                    & 0.385                                  & 0.386                                 & 0.581                                & 0.603          & 0.497          & 0.505          \\
        RaLD                                                      & \textbf{0.339}                         & \textbf{0.356}                        & \textbf{0.576}                       & \textbf{0.515} & \textbf{0.488} & \textbf{0.450} \\
        \bottomrule \hhline{=======}
    \end{tabular}
    \caption{Performance comparison of different radar point cloud generation
    models.}
    \label{tab:mian-diffusion}
\end{table}
% Next, we present the end-to-end results of the proposed RaLD model in Table~\ref{tab:mian-diffusion}
% and the visualization results in Figure~\ref{fig-vis-result}. Results show that
% RaLD achieves the best performance across all scenes, outperforming the baseline
% methods by a significant margin. Specifically, RaLD achieves a maximum 11.9\%
% improvement in CD and 14.6\% improvement in EMD compared to the second-best method,
% SDDiff, in the Aspen Lab and Hallways scenes, respectively. As the Figure~\ref{fig-vis-result}
% shows, RaLD can generate high-quality 3D radar point clouds that preserve more
% high-frequency details compared to the baseline methods, while the baseline methods
% SDDiff tend to generate smoother point clouds with less details. The results
% indicate that the proposed RaLD model can effectively leverage the learned
% latent space of the autoencoder to generate high-quality 3D radar point clouds.

\begin{table}[htb]
    \centering
    \fontsize{9pt}{10pt}\selectfont
    \setlength{\tabcolsep}{1.6pt}
    \begin{tabular}{@{}ccccccccccc@{}}
        \hhline{=========} \toprule \multirow{2}{*}{\textbf{Variant}} & \multirow{2}{*}{\makecell[c]{\textbf{Radar}\\\textbf{Enc.}}} & \multirow{2}{*}{\makecell[c]{\textbf{CFAR}\\\textbf{Init}}} & \multicolumn{2}{c}{\textbf{Aspen Lab}} & \multicolumn{2}{c}{\textbf{Hallways}} & \multicolumn{2}{c}{\textbf{ARPG Lab}} \\
        \cmidrule(lr){4-5} \cmidrule(lr){6-7} \cmidrule(lr){8-9}      &                                                              &                                                             & CD↓                                    & EMD↓                                  & CD↓                                  & EMD↓           & CD↓            & EMD↓           \\
        \midrule (a)                                                  & w/o                                                          & w/                                                          & 0.596                                  & 0.638                                 & 0.723                                & 0.647          & 0.659          & 0.628          \\
        (b)                                                           & w/                                                           & w/o                                                         & 0.348                                  & 0.381                                 & 0.586                                & 0.545          & 0.535          & 0.547          \\
        (c)                                                           & w/                                                           & w/                                                          & \textbf{0.339}                         & \textbf{0.356}                        & \textbf{0.576}                       & \textbf{0.515} & \textbf{0.488} & \textbf{0.450} \\
        \bottomrule \hhline{=========}
    \end{tabular}
    \caption{Ablation study on radar encoder conditioning and decoder query
    initialization. “w/” and “w/o” indicate the presence and absence of each component,
    respectively. Variants (a)–(c) correspond to different combinations of these
    modules.}
    \label{tab:diffusion-ablation_table}
\end{table}

Next, we report the end-to-end generation results of the proposed RaLD model in Table~\ref{tab:mian-diffusion},
with qualitative examples shown in Figure~\ref{fig-vis-result}. RaLD consistently
outperforms baseline methods across all scenes, achieving up to 11.9\% and 14.6\%
improvements in CD and EMD, respectively, compared to the second-best method, SDDiff,
in the Aspen Lab and Hallways scenes.

As shown in Figure~\ref{fig-vis-result}, RaLD produces sharper and more detailed
3D radar point clouds, better preserving high-frequency structures, while baseline
method SDDiff tends to generate overly smooth outputs. For example, in the Aspen
Lab and Hallways scenes, RaLD successfully captures fine-grained structures such
as pillars and wall edges, whereas SDDiff produces smoother outputs that omit these
details. While SDDiff generates point clouds with a more continuous surface, it often
lacks the key structural features present in the space, which can be crucial for
downstream tasks like localization and mapping. In contrast, RaLD effectively
captures these features, demonstrating its ability to leverage the learned latent
space of the autoencoder to generate high-fidelity 3D radar point clouds.
% These results demonstrate that RaLD effectively leverages the learned
% latent space to generate high-fidelity 3D radar point clouds.

\subsubsection{Ablation Studies and Additional Results}
We conduct ablation studies to evaluate the impact of two key components in the RaLD
model: radar encoder conditioning and decoder query initialization. The results are
summarized in Table~\ref{tab:diffusion-ablation_table}. We compare three model
variants: (a) without radar encoder conditioning only, (b) with radar encoder
conditioning only, and (c) with both components enabled. In variant (a), the raw
radar spectrum is flattened and embedded with positional encodings to serve as the
conditioning input to the diffusion model. For variants that do not use CFAR
points for decoder query initialization, we randomly initialize the same number of
points as used in the CFAR-based setup.

The results demonstrate that variant (c), which combines both radar encoder
conditioning and decoder query initialization consistently achieves the best
performance across all scenes. The comparison between variants (a) and (c) shows
that the radar encoder conditioning significantly improves performance,
indicating that the learned representation effectively captures radar signal
characteristics and mitigates noise in the raw radar spectrum. Meanwhile, the
comparison between variants (b) and (c) reveals that decoder query
initialization further enhances performance, suggesting that initializing decoder
queries with CFAR points provides a strong spatial prior for generating high-quality
point clouds.

We further investigate how autoencoder design impacts \sysname{} performance. Specifically,
we compare four types of encoder query strategies under different coordinate
systems, as summarized in Table~\ref{tab:diffusion-different-ae}. Across all
scenes, the diffusion model using the hybrid query in the frustum-based occupancy
partitioning consistently achieves the best performance in terms of EMD. For the
CD metric, this configuration achieves the best result in the Aspen Lab scene and
performs comparably to the downsampled point query in the other scenes. These
results suggest that the hybrid query strategy within a frustum-aligned latent
space is more effective at capturing the underlying 3D structure, leading to
higher-quality point cloud generation. This also validates our design of order-invariant
latent encoding, where combining static and dynamic queries enables stable
supervision and better generalization for diffusion-based generation. While our method
does not consistently produce the lowest CD—often sensitive to density and outliers~\cite{wu2021density}—it
consistently improves EMD, reflecting better overall structure.

% \begin{table}[tb]
\begin{table}[htb]
    \centering
    \fontsize{9pt}{10pt}\selectfont
    \setlength{\tabcolsep}{4pt} % control hor spacing
    \begin{tabular}{@{}lccccc@{}}
        \hhline{======} \toprule \multirow{2}{*}{\textbf{Scene}} & \makecell[c]{\textbf{Encoder}\\\textbf{Query}} & Hybrid & Sample         & Static  & Hybrid         \\
        \cmidrule{2-6}                                           & \textbf{Occupancy}                             & Voxel  & Frustum        & Frustum & Frustum        \\
        \midrule \multirow{2}{*}{Aspen Lab}                      & CD↓                                            & 0.397  & 0.366          & 0.390   & \textbf{0.339} \\
                                                                 & EMD↓                                           & 0.519  & 0.422          & 0.412   & \textbf{0.356} \\
        \midrule \multirow{2}{*}{Hallways}                       & CD↓                                            & 0.695  & \textbf{0.562} & 0.633   & 0.576          \\
                                                                 & EMD↓                                           & 0.770  & 0.566          & 0.540   & \textbf{0.515} \\
        \midrule \multirow{2}{*}{ARPG Lab}                       & CD↓                                            & 0.609  & \textbf{0.475} & 0.564   & 0.488          \\
                                                                 & EMD↓                                           & 0.766  & 0.511          & 0.513   & \textbf{0.450} \\
        \bottomrule \hhline{======}
    \end{tabular}
    \caption{Performance comparison of different autoencoder queries under
    different coordinate systems.}
    \label{tab:diffusion-different-ae}
\end{table}
\section{Additional Experiment Results}

\subsection{Generalization to Unseen Scenarios}

\subsubsection{Unseen Indoor Scenarios}
We validate the generalization ability of our method by conducting experiments
on the unseen indoor scenarios. We use the indoor dataset in SDDiff~\cite{wang2025sddiff},
and evaluate the performance of our autoencoder without any fine-tuning on the
unseen indoor scenes. The quantitative results are shown in Table~\ref{tab:ae_unseen_indoor}.
The results show that the hybrid encoder with frustum-based occupancy
partitioning outperforms the static encoder and the voxel-based partitioning
ones. And the performance degradation is not significant compared to the seen indoor
scenarios, which indicates that our autoencoder can generalize well to unseen
indoor scenarios.

\begin{table}[ht]
    \centering
    \fontsize{9pt}{10pt}\selectfont
    \setlength{\tabcolsep}{4pt} % control hor spacing
    \begin{tabular}{@{}lccccc@{}}
        \hhline{======} \toprule \multirow{2}{*}{\textbf{Scene}} & \makecell[c]{\textbf{Encoder}\\\textbf{Query}} & \textbf{Hybrid} & \textbf{Sample} & \textbf{Static} & \textbf{Hybrid} \\
        \cmidrule{2-6}                                           & \textbf{Occupancy}                             & Voxel           & Frustum         & Frustum         & Frustum         \\
        \midrule \multirow{2}{*}{Classroom}                      & CD↓                                            & 0.105           & \textbf{0.064}  & 0.096           & 0.080           \\
                                                                 & EMD↓                                           & 0.121           & \textbf{0.066}  & 0.078           & 0.073           \\
        \bottomrule \hhline{======}
    \end{tabular}
    \caption{Performance comparison of different autoencoders.}
    \label{tab:ae_unseen_indoor}
\end{table}

We further evaluate the generalization ability of our radar-based 3D point cloud
generation model. We fine-tune the model on the indoor scene with 20 epochs. The
quantitative results are shown in Table~\ref{tab:diffusion_finetune_indoor}. The
results show that our method achieves performance in unseen indoor scenarios
comparable to that in seen scenarios, with only 20 epochs of fine-tuning.

\begin{table}[ht]
    \centering
    \fontsize{9pt}{10pt}\selectfont
    \setlength{\tabcolsep}{10pt}
    \begin{tabular}{lccc}
        \hhline{====} \toprule \multirow{2}{*}{\textbf{Model}} & \multicolumn{2}{c}{\textbf{Classroom}} &                \\
        \cmidrule(lr){2-3}                                     & CD↓                                    & EMD↓           \\
        \midrule OS-CFAR                                       & 1.127                                  & 0.849          \\
        RaLD                                                   & \textbf{0.356}                         & \textbf{0.366} \\
        \bottomrule \hhline{====}
    \end{tabular}
    \caption{Generalization performance of radar point cloud generation in
    unseen indoor scenarios.}
    \label{tab:diffusion_finetune_indoor}
\end{table}

\subsubsection{Unseen Outdoor Scenarios}
To further validate the generalization ability of our method, we conduct experiments
on the unseen outdoor scenarios of the ColoRadar dataset. We first provide the
quantitative results of our autoencoder without any fine-tuning on the unseen
outdoor scenes in Table~\ref{tab:ae_unseen_outdoor}.

It can be seen that all the autoencoders encounter performance degradation
compared to the seen indoor scenarios. Although it still performs the worst among
all invariant encoders, the voxel-based partitioning autoencoder shows a
significant performance drop compared to indoor scenarios, indicating that voxel-based partitioning is particularly ill-suited for outdoor environments. Within the
frustum-based partitioning autoencoders, the hybrid encoder still outperforms the
static encoder, which shows that the hybrid encoder also has better
generalization ability to unseen outdoor scenarios. Notably, the performance gap
between the hybrid encoder and the downsampled point encoder further widens in
outdoor settings. We speculate that this is due to the increased sparsity of outdoor
scenes, which lies outside the distribution of the training data. We anticipate that
the hybrid encoder will perform even better in outdoor environments when such scenes
are included during training.

\begin{table}[htb]
    \centering
    \fontsize{9pt}{10pt}\selectfont
    \setlength{\tabcolsep}{4pt} % control hor spacing
    \begin{tabular}{@{}lccccc@{}}
        \hhline{======} \toprule \multirow{2}{*}{\textbf{Scene}} & \makecell[c]{\textbf{Encoder}\\\textbf{Query}} & \textbf{Hybrid} & \textbf{Sample} & \textbf{Static} & \textbf{Hybrid} \\
        \cmidrule{2-6}                                           & \textbf{Occupancy}                             & Voxel           & Frustum         & Frustum         & Frustum         \\
        \midrule \multirow{2}{*}{Longboard}                      & CD↓                                            & 0.369           & \textbf{0.113}  & 0.201           & 0.178           \\
                                                                 & EMD↓                                           & 0.466           & \textbf{0.112}  & 0.198           & 0.177           \\
        \midrule \multirow{2}{*}{Courtyard}                      & CD↓                                            & 0.390           & \textbf{0.114}  & 0.181           & 0.160           \\
                                                                 & EMD↓                                           & 0.160           & \textbf{0.115}  & 0.184           & 0.164           \\
        \bottomrule \hhline{======}
    \end{tabular}
    \caption{Performance comparison of different autoencoders.}
    \label{tab:ae_unseen_outdoor}
\end{table}

Then We fine-tune the model on two outdoor scenes, Longboard and Courtyard, also
with 20 epochs. The quantitative results are shown in Table~\ref{tab:diffusion_finetune_outdoor}.
Although the absolute performance remains lower than the performance in both
seen and unseen indoor scenarios, the fine-tuned model still achieves substantial
improvements over the OS-CFAR baseline. We attribute this performance gap to the
sparsity and complexity of outdoor point clouds, which make the generation task more
difficult, as the significant performance degradation is also observed in OS-CFAR
baseline. Additionally, the reduced autoencoder quality in these scenes may further limit the generation performance.

\begin{table}[ht]
    \centering
    \fontsize{9pt}{10pt}\selectfont
    \setlength{\tabcolsep}{10pt}
    \begin{tabular}{lcccc}
        \hhline{=====} \toprule \multirow{2}{*}{\textbf{Model}} & \multicolumn{2}{c}{\textbf{Longboard}} & \multicolumn{2}{c}{\textbf{Courtyard}} \\
        \cmidrule(lr){2-3} \cmidrule(lr){4-5}                   & CD↓                                    & EMD↓                                  & CD↓            & EMD↓           \\
        \midrule OS-CFAR                                        & 3.565                                  & 4.256                                 & 1.372          & 1.541          \\
        RaLD                                                    & \textbf{0.866}                         & \textbf{1.109}                        & \textbf{0.720} & \textbf{0.796} \\
        \bottomrule \hhline{=====}
    \end{tabular}
    \caption{Generalization performance of radar point cloud generation in
    unseen outdoor scenarios.}
    \label{tab:diffusion_finetune_outdoor}
\end{table}

\subsection{Model Scalability}
We also evaluate the scalability of our model by changing the number of DiT blocks
in the diffusion model. The results are shown in Table~\ref{tab:diffusion_scalability}.
It can be seen that the performance improves, at most cases, as the number of DiT~\cite{peebles2023scalable}
blocks increases, which indicates that the model can benefit from deeper
architectures. However, the improvement is not significant, and the performance saturates
when the number of DiT blocks reaches 24. To thoroughly explore the upper capacity
of our model, we adopt the depth of 24 as the default setting in our experiments.

\begin{table}[htb]
    \centering
    \fontsize{9pt}{10pt}\selectfont
    \setlength{\tabcolsep}{1.6pt}
    \begin{tabular}{@{}ccccccccc@{}}
        \hhline{========} \toprule \multirow{2}{*}{\textbf{Scale}} & \multirow{2}{*}{\makecell[c]{\textbf{Param (M)}}} & \multicolumn{2}{c}{\textbf{Aspen Lab}} & \multicolumn{2}{c}{\textbf{Hallways}} & \multicolumn{2}{c}{\textbf{ARPG Lab}} \\
        \cmidrule(lr){3-4} \cmidrule(lr){5-6} \cmidrule(lr){7-8}   &                                                   & CD↓                                    & EMD↓                                  & CD↓                                  & EMD↓           & CD↓            & EMD↓           \\
        \midrule Depth = 12                                        & 101.87                                            & 0.367                                  & 0.388                                 & 0.591                                & 0.518          & 0.493          & \textbf{0.448} \\
        Depth = 18                                                 & 142.82                                            & 0.361                                  & 0.383                                 & 0.590                                & 0.520          & 0.500          & 0.449          \\
        Depth = 24                                                 & 183.77                                            & \textbf{0.339}                         & \textbf{0.356}                        & \textbf{0.576}                       & \textbf{0.515} & \textbf{0.489} & 0.450          \\
        \bottomrule \hhline{========}
    \end{tabular}
    \caption{Scalability evaluation of the generation model with different
    parameter sizes.}
    \label{tab:diffusion_scalability}
\end{table}

\begin{figure*}[htb]
% \begin{figure*}[tb]
    \centering
    \includegraphics[width=\textwidth]{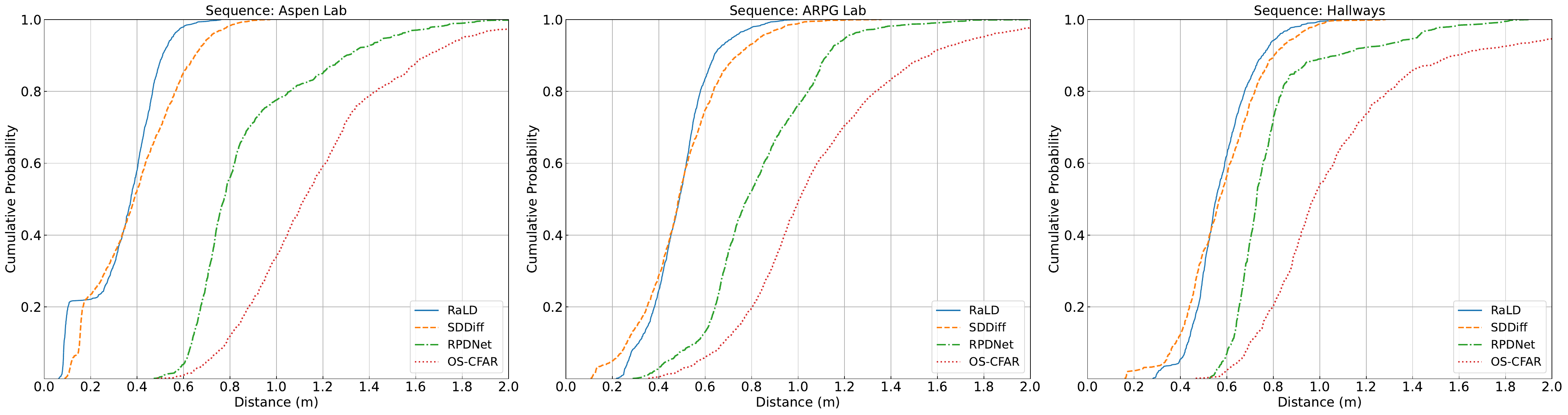}
    \caption{Cumulative distribution function (CDF) of CD for different methods
    cross scenes.}
    \label{fig:cd_cdf}
\end{figure*}
\begin{figure*}[htb]
% \begin{figure*}[tb]
    \centering
    \includegraphics[width=\textwidth]{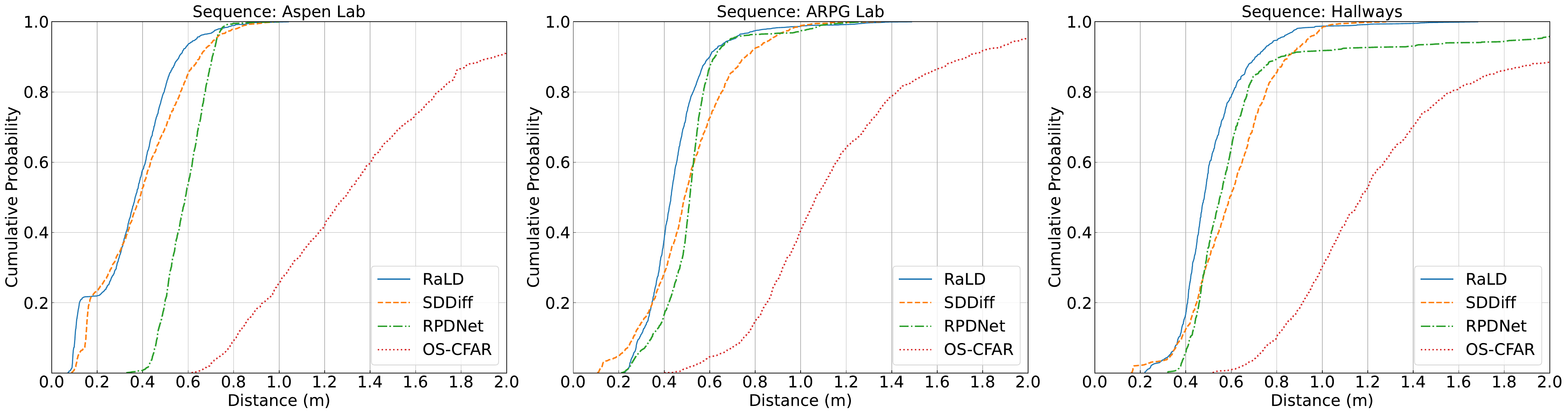}
    \caption{Cumulative distribution function (CDF) of EMD for different methods
    cross scenes.}
    \label{fig:emd_cdf}
\end{figure*}

\subsection{Additional Results on Generation Performance}
We provide additional results from the main radar 3D point cloud generation experiment,
including detailed cumulative distribution function (CDF) curves for CD and EMD.
These are shown in Figure~\ref{fig:cd_cdf} and Figure~\ref{fig:emd_cdf},
respectively.
\section{Conclusion}
We propose a latent diffusion framework that generates high-resolution 3D point
clouds from noisy radar spectrum. By combining a frustum-based autoencoder, order-invariant
encoding, and radar spectrum guidance, our method effectively reconstructs
detailed scene geometry, advancing radar-based 3D perception in challenging environments.
% \input{AnonymousSubmission/LaTeX/supp.tex}
% \newpage
\bibliography{RaLD_ref}
\end{document}